\documentclass[11pt,a4paper]{article}
\usepackage[hyperref]{emnlp-ijcnlp-2019}
\usepackage{times}
\usepackage{latexsym}
\usepackage{amsmath}
\usepackage{CJK}
\usepackage{indentfirst}
\usepackage{graphicx}

\usepackage{url}

\aclfinalcopy 

\title{Generating Highly Relevant Questions}

\author{Jiazuo Qiu \and Deyi Xiong\thanks{\ \ Corresponding author} \\
        School of Computer Science and Technology, Soochow University, Suzhou, China \\
        {\tt qjzhzw@163.com}; {\tt dyxiong@suda.edu.cn} \\}

\date{}

\begin{document}
\maketitle
\begin{abstract}
The neural seq2seq based question generation (QG) is prone to generating generic and undiversified questions that are poorly relevant to the given passage and target answer. In this paper, we propose two methods to address the issue. (1) By a partial copy mechanism, we prioritize words that are morphologically close to words in the input passage when generating questions; (2) By a QA-based reranker, from the n-best list of question candidates, we select questions that are preferred by both the QA and QG model. Experiments and analyses demonstrate that the proposed two methods substantially improve the relevance of generated questions to passages and answers.
\end{abstract}

\section{Introduction}
\label{sec:length}

Question generation is to generate a valid and fluent question according to a given passage and the target answer. In general, the answer is a span of words in the passage. QG can be used in many scenarios, such as automatical tutoring  systems, improving the performance of QA models and enabling chatbots to lead a conversation.

In the early days, QG has been tackled mainly via rule-based approaches~\cite{Mitkov:03, Heilman:10}. These methods rely on many handcrafted rules and templates. Constructing such rules is time-consuming and it is difficult to adapt them to other domains.

Very recently, neural networks have been used for QG. Specifically, the encoder-decoder seq2seq model ~\cite{Du:17, Zhou:17, Song:18} is used to encode a passage and generate a question corresponding to the answer.

Such end-to-end neural models are able to generate better questions than traditional rule-based approaches. However, one issue with the current neural models is that the generated questions are not quite relevant to the corresponding passages and target answers. In other words, the neural QG models tend to generate generic questions (e.g., ``what is the name of...?'').

In this paper, we propose two methods to deal with this low-relevance issue for QG. First, we present a partial copy method to enhance the existing copy mechanism so that the QG model can not only copy a word as an entire unit from the passage to the generated question but also copy a part of a word (e.g., ``start'' in ``started'') to generate a new morphological form of the word in the question. The fine-grained partial copy mechanism enables the QG model to copy morphologically changed words from the passage, increasing the relevance of the generated question to the passage by sharing more words in different forms. Second, we propose a QA-based reranker to rerank QG results. Particularly, we use a neural QA model to evaluate the quality of generated questions, and rerank them according to the QA model scores. Normally, generic questions get low scores from the neural QA model, hence the reranker is able to select non-generic and highly relevant questions.

While alleviating the low-relevance issue, the proposed two methods alone and their combination have also improved the quality of generated questions in terms of both BLEU and METEOR in our experiments.

\section{Methods}
\label{sec:length}

\subsection{The Partial Copy Mechanism}

The conventional copy mechanism ~\cite{Gu:16, See:17} can allow the decoder to copy a word from the input passage to the generated question. However, such copy mechanism works at the word level. It cannot copy a part of a word to reproduce an appropriate form of the word in the generated question. In other words, it cannot copy words with morphological changes. However, morphological changes frequently happen when we transform a passage into a question as grammatical functions of some words (e.g., verbs, nouns, adjectives, etc) change.

Let's consider the following example:
\begin{quote}
Passage: Teaching started in 1794. \\
Question: When did teaching start?
\end{quote}
The morphological variants ``start'', ``starts'' of ``started'' in the passage may be used in the generated question.

In order to encourage the decoder to copy the inflected forms of a word from the passage to the generated question, we propose a partial copy mechanism that measures the character overlap rate of an original word in the passage with its copied form in the question. We first detect the overlapped subsequence of characters between words $w_{1}$ and $w_{2}$ according to the longest common subsequence (LCS) between them. For example, ``start'' and ``started'' have LCS ``start''. Then we calculate the overlap rate $C$ between $w_{1}$ and $w_{2}$ as follows.
\begin{equation}
C=\frac{|LCS|*2}{|w_{1}|+|w_{2}|}
\end{equation} 
According to this formula, the overlap rate between ``start'' and ``started'' is 0.71. The value range of $C$ is $\in [0,1]$. Full-word copy is of course encouraged as the value of $C$ is 1.

One problem with $C$ is that many unrelated words also have LCS. For example, ``a'' is the LCS of ``append'' and ``start''. The overlap rate of the two words is 0.18 instead of 0. To ensure the effectiveness of the method, we compute the final overlap rate with a threshold $\gamma$ :
\begin{equation}
C = 
\begin{cases} 
C,  & \mbox{if } C >= \gamma \\
0,  & \mbox{if } C < \gamma
\end{cases}
\end{equation}

Whenever we generate a word in the question, we find its corresponding word in the passage with the highest attention weight. We then calculate the overlap rate $C$ between the two words and use $C$ to re-adjust the probability of the generated word as follows:
\begin{equation}
P_{adj}=P*(1+\lambda_{1}*C)
\end{equation}
where $P$ is the original generation probability output by the decoder, $\lambda_{1}$ is a hyperparameter whose range is [0, $+\infty$). We normalize all these re-adjusted probabilities to get the final probability distribution.

\begin{table*}[t!]
\centering
\begin{tabular}{llllll}
  models & BLEU-1 & BLEU-2 & BLEU-3 & BLEU-4 & METEOR \\
  \hline
  \citet{Du:17} & 43.09 & 25.96 & 17.50 & 12.28 & 16.62 \\
  \citet{Song:18} (reported in paper) & -- & -- & -- & 13.98 & 18.77 \\
  \citet{Song:18} (our re-running) & 42.15 & 27.21 & 19.53 & 14.56 & 19.15 \\
  \hline
  partial copy ($\lambda_{1}$=0.5) & 44.13 & 28.29 & 20.17 & 14.95 & 19.81 \\
  partial copy ($\lambda_{1}$=1) & 44.33 & 28.34 & 20.17 & 14.95 & 20.00 \\
  partial copy ($\lambda_{1}$=2) & 42.34 & 26.95 & 19.15 & 14.16 & {\bf 20.16} \\
  QA-based reranking ($\lambda_{2}$=0.2) & 42.64 & 27.57 & 19.77 & 14.72 & 19.43 \\
  QA-based reranking ($\lambda_{2}$=0.5) & 42.50 & 27.42 & 19.62 & 14.56 & 19.38 \\
  QA-based reranking ($\lambda_{2}$=0.8) & 42.43 & 27.34 & 19.54 & 14.48 & 19.34 \\
  partial copy + QA-based reranking & {\bf 44.61} & {\bf 28.78} & {\bf 20.59} & {\bf 15.29} & 20.13 \\
\end{tabular}
\caption{Experiment results on the test set.}
\end{table*}

\subsection{The QA-Based Reranking}

Since we use the beam search algorithm in the neural QG decoder, we can generate multiple question candidates. We find that the generated question with the highest probability according to the baseline neural QG model is not always the best question.

Partially inspired by \citet{Li:16}, we propose a QA-based reranker to rerank the  n-best questions generated by the baseline decoder.

In general, the task of the QG model is to estimate the probability $P(q|p,a)$ of a generated question $q$ given the passage $p$ and target answer $a$. In the QA-based reranker, we re-estimate the quality of a candidate question by calculating the probability of the target answer $a$ given the passage and the generated question $q$, i.e., $P(a|p, q)$.

In theory, we can combine the two probabilities to rerank generated questions. But in practice, we take a more intuitive and straightforward way to use the $F_{1}$ score of a predicted answer by the trained QA model according to the generated question. The $F_{1}$ score is calculated at the character level by comparing the generated answer and gold answer (considering the answers as a set of characters). The idea behind this is that a good question allows the QA model to easily find an answer close to the ground truth answer.

The score used to rerank a question candidate is therefore computed as:
\begin{equation}
score=(1-\lambda_{2})*score_{1}+\lambda_{2}*score_{2}
\end{equation}
where $score_{1}$ is the log probability of the candidate question estimated by the baseline QG model, $score_{2}$ is the $F_{1}$ score of the predicted answer by the QA model, and $\lambda_{2}$ is a hyperparameter whose range is [0, 1].

\section{Experiments}
\label{sec:length}

\subsection{Datasets}

Following previous work, we conducted our experiments on SQuAD \cite{Rajpurkar:16}, a QA dataset which can also be used for QG. The dataset contains 536 articles and over 100k questions. Since the test set is unavailable, \citet{Du:17} randomly divide the raw dataset into train/dev/test set. In our experiments, we used the same data split as \citet{Du:17}.

\subsection{Baseline and Settings}

Our baseline is based on the QG model proposed by ~\citet{Song:18}. To be specific, it is a seq2seq model with attention and copy mechanism. The model consists of two encoders and a decoder. The two encoders encode a passage = ($p_{1}$,...,$p_{M}$) and an answer = ($a_{1}$,...,$a_{N}$) respectively. Additionally, multi-perspective matching strategies are used to combine the two encoders. With information from the encoders, the decoder generates a question word by word.

We retained the same values for most hyperparameters in our experiments as the baseline system \cite{Song:18}. We used Glove \cite{Pennington:14} to initialize the word embeddings and trained the model for 10 epochs. Copy and coverage mechanism \cite{See:17} were included while additional lexical features (POS, NER) were not. We used adam \cite{Kingma:15} as the optimizer during training. The beam size was set to 20 for the decoder.

In the experiments of the partial copy mechanism, we set the threshold $\gamma$ to 0.7. Three values (0.5, 1 and 2) were tried for $\lambda_{1}$.

In the experiments of the QA-based reranking, we used the SAN model \cite{Liu:18} as the QA model. 0.2, 0.5 and 0.8 were tried for $\lambda_{2}$.

We used automatic evaluation metrics: BLEU-1, BLEU-2, BLEU-3, BLEU-4 \cite{Papineni:02} and METEOR \cite{Denkowski:14}. The calculation script was provided by \citet{Du:17}.

\begin{table*}
\centering
\begin{tabular}{lll}
  Question templates & \citet{Song:18} & Our model \\
  \hline
  What is/was the name of ...? & 8,180/13,740 & 5,618/9,990 \\
  \hline
  What type of ...? & 4,942 & 3,805 \\
  \hline
  What is/was another name ...? & 2,731/26 & 57/3 \\
  \hline
  What is/was the total ...? & 470/794 & 101/62 \\
  \hline
  What is/was it ...? & 251/57 & 111/32 \\
\end{tabular}
\caption{Frequency of generic questions generated by the baseline and our methods. }
\end{table*}

\begin{table}
\centering
\small
\begin{tabular}{p{0.9\columnwidth}}
  \hline
  {\bf Passage}: in the polytechnic sector : wellington polytechnic amalgamated with massey university . \\
  \hline
  {\bf Answer}: wellington polytechnic \\
  \hline
  {\bf Question}: what school did massey university combine with ? \\
  \hline
  {\bf Baseline}: what is the name of the polytechnic sector in the polytechnic ? \\
  \hline
  {\bf Partial copy}: who amalgamated with massey university in the polytechnic sector ? \\
  \hline
  ~\\
  ~\\
  \hline
  {\bf Passage}: in practice , catholic services in all provinces were quickly forbidden , and the reformed church became the `` public '' or `` privileged '' church in the republic . \\
  \hline
  {\bf Answer}: catholic services \\
  \hline
  {\bf Question}: what was forbidden in all provinces ? \\
  \hline
  {\bf Baseline}: what was the `` public '' church in the republic ? \\
  \hline
  {\bf QA-based reranking}: what was forbidden in all provinces in the republic ? \\
  \hline
\end{tabular}
\caption{QG examples.}
\end{table}

\subsection{Results}

Experiment results are shown in Table 1, from which we observe that our methods substantially improves the baseline in terms of all evaluation metrics.

The combination of the proposed two methods ($\lambda_1$ = 1, $\lambda_2$ = 0.2) achieve the best performance, gaining 0.73 BLEU-4 and nearly 1 METEOR point of improvements over the baseline (obtained by re-running the source code of the baseline, higher than the results reported in the paper \cite{Song:18}). The proposed partial copy mechanism alone obtain substantial improvements over the baseline, especially in terms of BLEU-1 and BLEU-2. This is because this method is able to help the decoder copy morphologically changed words from passages. The application of the QA-based reranking obtain further improvements over the partial copy mechanism, indicating that the two methods are complementary to each other.

\section{Analysis}
\label{sec:length}

{\bf Partial copy}: We use the method to enhance the existing copy mechanism, making generated questions more relevant to passages and target answers. The average proportion of words (or their other morphological forms) that are copied from the passages in generated questions increases from 75.49\% (the baseline) to 78.74\%. Under such a mechanism, generic questions such as ``what is the name of...?'' will be penalized by our method as the overlap rate between words in these generic questions and those in passages is low. On the contrary, questions with higher overlap rate and therefore higher relevance to the input passage are rewarded by the new copy mechanism. In order to testify this hypothesis, we counted the numbers of such generic questions generated by the baseline and our method, which are shown in Table 2. It is clearly seen that the number of generic questions is significantly decreased after the partial copy mechanism is used.

In the first example displayed in Table 3,  the phrase ``what school'' is difficult to be generated as it does not appear in the input passage. The baseline model generates the generic question ``what is the name of...?'', which is not relevant to the passage and target answer. Such generic questions are generated because the templates of these questions occur frequently in the training data. The trained seq2seq model is prone to generating these ``safe'' questions, similar to the undiversified response generation in seq2seq-based dialogue model \cite{Li:16}. In contrast, our model is able to generate a more relevant question including a rare word ``amalgamated'' as the word has a high overlap rate.

{\bf QA-based reranking}: In our experiments, a total of 2,099 questions were reranked. Among them, 1,117 examples achieve a higher BLEU-4 score after reranking, while only 821 examples have a lower BLEU-4 after reranking.

In the second example of Table 3, the question with the highest score generated by the baseline is ``what church'', while the ground truth question is asking ``what was forbidden''. Since the generated questions to be reranked are different to each other, the QA model naturally finds different answers to these questions. For ``what church'', the QA model detected ``reformed church'' as the answer while for ``what was forbidden'', the QA model correctly detected the target answer ``catholic services''. Therefore, the QA-based reranker is able to find the answer-relevant questions.

\section{Related Work}
\label{sec:length}

The neural QG is an emerging task. Unlike the extractive QA, most neural QG models are generative. \citet{Du:17} pioneer the neural QG by proposing neural seq2seq models to deal with the task. Unfortunately, they do not use the target answer for QG. At the same time, \citet{Zhou:17} present a similar model for QG. They use answer position embeddings to represent target answers and explore a variety of lexical features.

After that, many QG studies have been conducted on the basis of the widely-used seq2seq architecture together with the attention and copy mechanism. \citet{Song:18} propose two encoders for both the passage and the target answer. \citet{Du:18} employ coreferences as an additional feature. \citet{Kim:19} propose a model of answer separation. \citet{Yuan:17} and \citet{Kumar:18} adopt reinforcement learning to optimize the generation process.

QA and QG are closely related to each other. \citet{Tang:17} treat QA and QG as dual tasks, and many other studies use QG to enhance QA or jointly learn QG and QA \cite{Duan:17, Wang:17, Sachan:18}.

\section{Conclusion}
\label{sec:length}

In this paper, we have presented two methods to improve the relevance of generated questions to the given passage and target answer. Experiments and analyses on SQuAD show that both the partial copy mechanism and QA-based reranking improve the relevance of generated questions in terms of both BLEU and METEOR. 

\section*{Acknowledgements}
\label{sec:length}

The present research was supported by the National Natural Science Foundation of China (Grant No. 61622209). We would like to thank the anonymous reviewers for their insightful comments.

\bibliography{emnlp-ijcnlp-2019}
\bibliographystyle{acl_natbib}

\end{document}